\documentclass[conference]{IEEEtran}
\IEEEoverridecommandlockouts
\usepackage{cite}
\usepackage{amsmath,amssymb,amsfonts}%数学公式
\usepackage{algorithmic}
\usepackage{graphicx}%图片支持
\usepackage{textcomp}
\usepackage{xcolor}
\usepackage{booktabs} % 优化表格线宽 (\toprule, \midrule, \bottomrule)
\usepackage[table]{xcolor} % 颜色支持
\def\BibTeX{{\rm B\kern-.05em{\sc i\kern-.025em b}\kern-.08em
    T\kern-.1667em\lower.7ex\hbox{E}\kern-.125emX}}
\begin{document}
% --- 标题 (Title) ---
\title{KidVis: Do Multimodal Large Language Models Possess the Visual  Perceptual Capabilities of a 6-Year-Old?}

%\author{Anonymous ICME submission}

\author{
    \IEEEauthorblockN{
        Xianfeng Wang\IEEEauthorrefmark{1}, 
        Kaiwei Zhang\IEEEauthorrefmark{2}, 
        Qi Jia\IEEEauthorrefmark{2},
        Zijian Chen\IEEEauthorrefmark{1},
        Guangtao Zhai\IEEEauthorrefmark{1},
        Xiongkuo Min\IEEEauthorrefmark{1}
    }
    \IEEEauthorblockA{\IEEEauthorrefmark{1}Shanghai Jiao Tong University}
    \IEEEauthorblockA{\IEEEauthorrefmark{2}Shanghai AI Laboratory}
}

\maketitle

\begin{abstract}
While Multimodal Large Language Models (MLLMs) have demonstrated impressive proficiency in high-level reasoning tasks, such as complex diagrammatic interpretation, it remains an open question whether they possess the fundamental visual primitives comparable to human intuition. To investigate this, we introduce KidVis, a novel benchmark grounded in the theory of human visual development. KidVis deconstructs visual intelligence into six atomic capabilities—Concentration, Tracking, Discrimination, Memory, Spatial, and Closure-already possessed by 6-7 year old children, comprising 10 categories of low-semantic-dependent visual tasks. Evaluating 20 state-of-the-art MLLMs against a human physiological baseline reveals a stark performance disparity. Results indicate that while human children achieve a near-perfect average score of 95.32, the state-of-the-art GPT-5 attains only 67.33. Crucially, we observe a "Scaling Law Paradox": simply increasing model parameters fails to yield linear improvements in these foundational visual capabilities. This study confirms that current MLLMs, despite their reasoning prowess, lack the essential physiological perceptual primitives required for generalized visual intelligence.
\end{abstract}


\section{Introduction}
\label{sec:intro}

With the rapid evolution of Multimodal Large Language Models (MLLMs), artificial intelligence appears to be approaching the frontier of Artificial General Intelligence (AGI)\cite{ref01,ref02,ref39}. Contemporary MLLMs demonstrate expert-level proficiency not only in fine-grained image descriptions and open-domain Visual Question Answering (VQA) but also in high-level cognitive tasks, such as parsing complex chart logic, assisting in medical diagnosis, and performing spatiotemporal planning as embodied agents\cite{ref17,ref19,ref20}. However, a fundamental paradox underlies these advancements: while MLLMs can articulate the philosophical connotations of art or generate code from sketches via Chain-of-Thought, their robustness in elementary perceptual tasks remains surprisingly fragile\cite{ref15,ref18,ref40}. This phenomenon echoes Moravec’s Paradox: high-level semantic reasoning has become computationally accessible, yet sensorimotor tasks that are instinctive to human children remain significant obstacles for machines. This raises a critical question: Do current large-scale models truly perceive physical reality, or do they merely mimic perceptual behaviors driven by high-dimensional statistical correlations?

\begin{figure}[htbp]
\centerline{\includegraphics[width=\columnwidth]{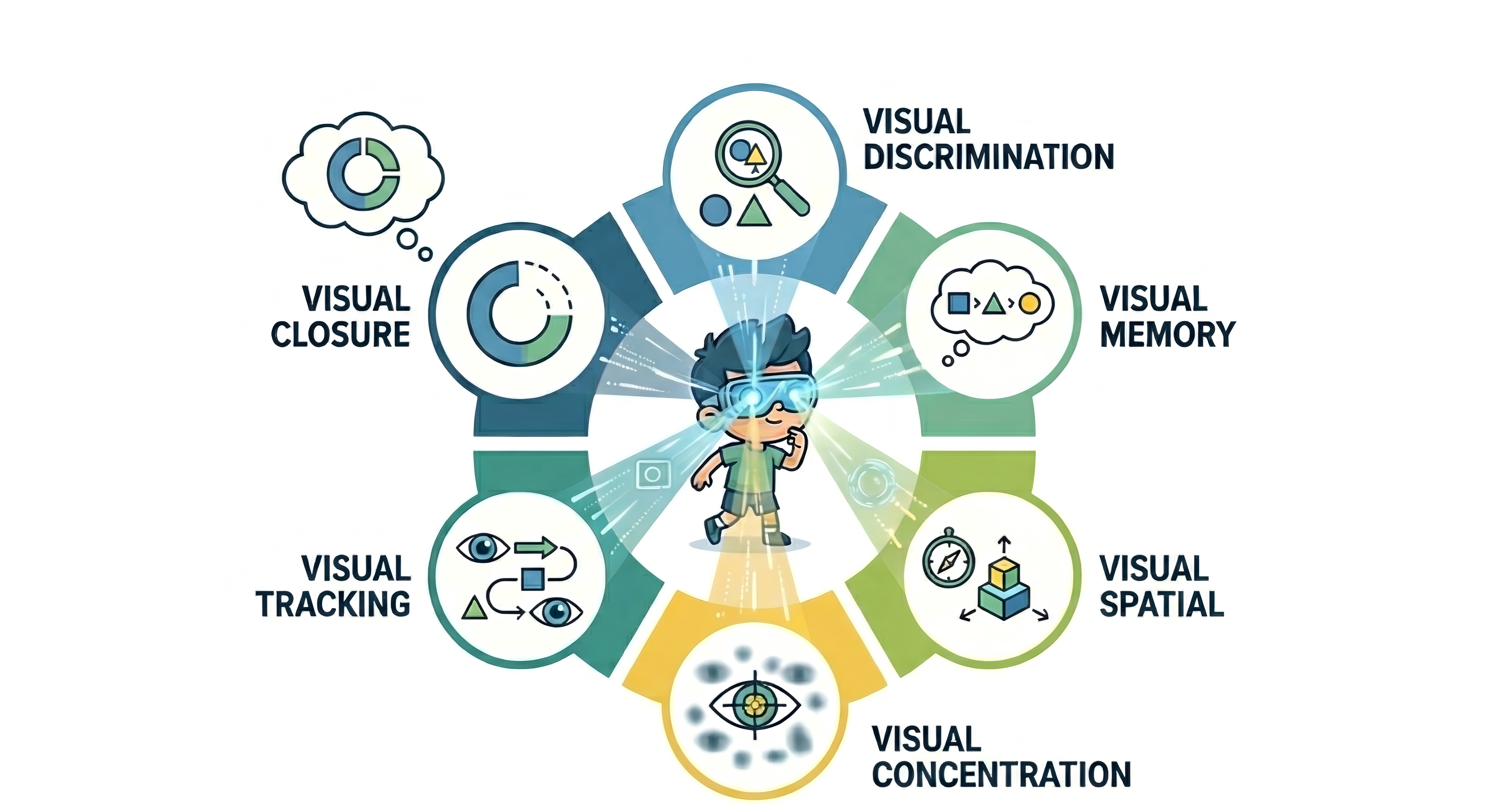}}
\caption{The scheme of visual 
perceptual capabilities in KidVis benchmark.}
\label{fig01}
\end{figure}

To investigate these limitations, substantial research efforts have studied the failure modes of MLLMs. For instance, studies on Object Hallucination attribute the generation of non-existent objects to an over-reliance on linguistic priors learned during pre-training, rather than faithful visual grounding\cite{ref05,ref06}. Similarly, experiments reveal that models exhibit severe biases in basic recognition tasks when visual stimuli conflict with inherent knowledge bases\cite{ref15}. While domain-specific benchmarks\cite{ref07} and comprehensive evaluation frameworks\cite{ref08,ref14} have exposed structural bottlenecks in mathematical reasoning and general multimodal understanding, they primarily focus on adult-level high-level semantic reasoning or provide fragmented diagnostic perspectives on specific symptoms. Crucially, existing works often fail to systematically decompose vision into constitutive cognitive primitives or benchmark these deficits against the foundational stages of human visual development.

In this study, we hypothesize that MLLM failures in complex scenarios stem from deficits in core visual primitives—capabilities that are typically fully developed in 6-7-year-old children. Drawing on cognitive development theories\cite{ref10,ref13}, we categorize these capabilities into six dimensions: Discrimination, Memory, Concentration, Tracking, Spatial, and Closure, as illustrated in Fig.~\ref{fig01}. To test this, we introduce KidVis, a benchmark comprising ten categories of "child-level" visual tasks. Unlike existing datasets rich in semantic context, these tasks serve as "diagnostic baselines": they minimize the interference of motor skills and semantic reasoning, strictly demanding precise visual processing mechanisms. We conduct an extensive zero-shot evaluation of 20 mainstream MLLMs, encompassing both proprietary models and open-source series, comparing them against a physiological baseline established by 6-7-year-old children.Our key findings are:
\begin{itemize}
\item Validation of Moravec's Paradox in Vision: While models excel in high-level reasoning, a substantial performance gap exists between MLLMs and the human baseline on elementary visual tasks, confirming that simple human instincts remain a significant challenge for Artificial Intelligence. 
\item Concept is superior to perception: Qualitative analysis reveals a systemic bias where MLLMs prioritize learned semantic priors and superficial texture consistency over rigorous geometric verification. Models frequently hallucinate biologically correct but visually false details or rely on color matching instead of boundary topology, indicating an inability to ground high-level concepts in pixel-level physical reality.
\item The Scaling Law Paradox: We observe a counter-intuitive phenomenon where increasing model parameters fails to yield linear improvements in foundational visual perception. This suggests that relying solely on model scaling is insufficient to endow MLLMs with true physical perception.
\end{itemize}

\section{KidVis Benchmark}

\subsection{Classification of Visual Capabilities}

To strictly assess the foundational visual perceptual capabilities of MLLMs, we move beyond high-level semantic reasoning tasks\cite{AIBench,zhang2025large}. Our framework is deeply rooted in the seminal cognitive factor analysis of visual perception. Specifically, we draw on Carroll’s Three-Stratum Theory\cite{ref10}, which theoretically deconstructs human visual behavior into distinct cognitive factors, including Visualization, Spatial Relations, Closure Speed, Flexibility of Closure and Perceptual Speed. Furthermore, to tailor this framework for disembodied MLLMs, we refer to standardized developmental protocols like the Motor-Free Visual Perceptual Test (MVPT-3)\cite{ref13} and Developmental Test of Visual Perception (DTVP-2)\cite{ref11}. While classical frameworks often confuse perception with motor skills (e.g., Eye-Hand Coordination in DTVP-2), MVPT-3 explicitly isolates motor-free components, focusing solely on the brain's processing of visual signals. By synthesizing these mature classification systems originally designed for human cognition, we establish a classification of six atomic visual primitives, namely Concentration, Tracking, Discrimination, Memory, Spatial, and Closure, grounded in developmental psychology.
\paragraph{Visual Concentration (VC)} The capabilities to sustain visual attention on specific stimuli while actively suppressing irrelevant objects from complex backgrounds\cite{ref27}. This serves as the prerequisite for valid visual processing.
\paragraph{Visual Tracking (VT)} The oculomotor ability to execute smooth pursuit along a path or trajectory, essential for maintaining topological continuity in reading and motion perception\cite{ref27}.
\paragraph{Visual Discrimination (VD)} The ability to distinguish fine-grained features, such as subtle variations in color, shape, position or texture, to categorize objects distinctively\cite{ref28}.
\paragraph{Visual Memory (VM)} Includes Visual Short-term Memory for buffering transient visual details (sequence, location) to support sequential reasoning\cite{ref29}, and Long-term Memory for retrieving encoded visual prototypes \cite{ref31,ref32}.
\paragraph{Visual Spatial (VS)} Entails perceiving spatial relationships (e.g., orientation, relative position) and performing mental manipulation (e.g., rotation) within a defined coordinate system\cite{ref29}.
\paragraph{Visual Closure (VCl)} The ability to identify an unknown visual object when only presented with a visual stimulus that is obscure, disconnected, incomplete, or vague\cite{ref29}.

\subsection{Task Instantiation}

\begin{table}[t]
\centering
% 如果您之前修改了caption设置，这里会自动应用
\caption{
  \parbox{\linewidth}{%  <-- 创建一个占满行宽的盒子
    \normalfont          %  <-- 1. 恢复正常字体（取消大写）
    \centering
    Mapping of tasks to their target visual capabilities.
  }
}

% 使用 \linewidth 确保表格适应单栏宽度
\setlength{\tabcolsep}{2mm}
% \resizebox{\linewidth}{!}{%
\begin{tabular}{l|cc}
\toprule
\textbf{Task Name} & \textbf{Target Cap. 1} & \textbf{Target Cap. 2} \\
\midrule
Body Part Counting & VD & VM \\
Clock Reading & VS & VD \\
Complex Scene Counting & VC & VD \\
Hidden Figures & VT & VCl \\
Schulte Grid & VC & VT \\
Spatial Orientation & VS & VD \\
Visual Completion & VM & VCl \\
Visual Reasoning & VD & VM \\
Jigsaw Assembly & VCl & VS \\
Path Tracing & VT & VC \\
\bottomrule
\end{tabular}%
% }
\label{tab01}
\end{table}

\begin{figure*}[htbp]
\centerline{\includegraphics[width=\textwidth]{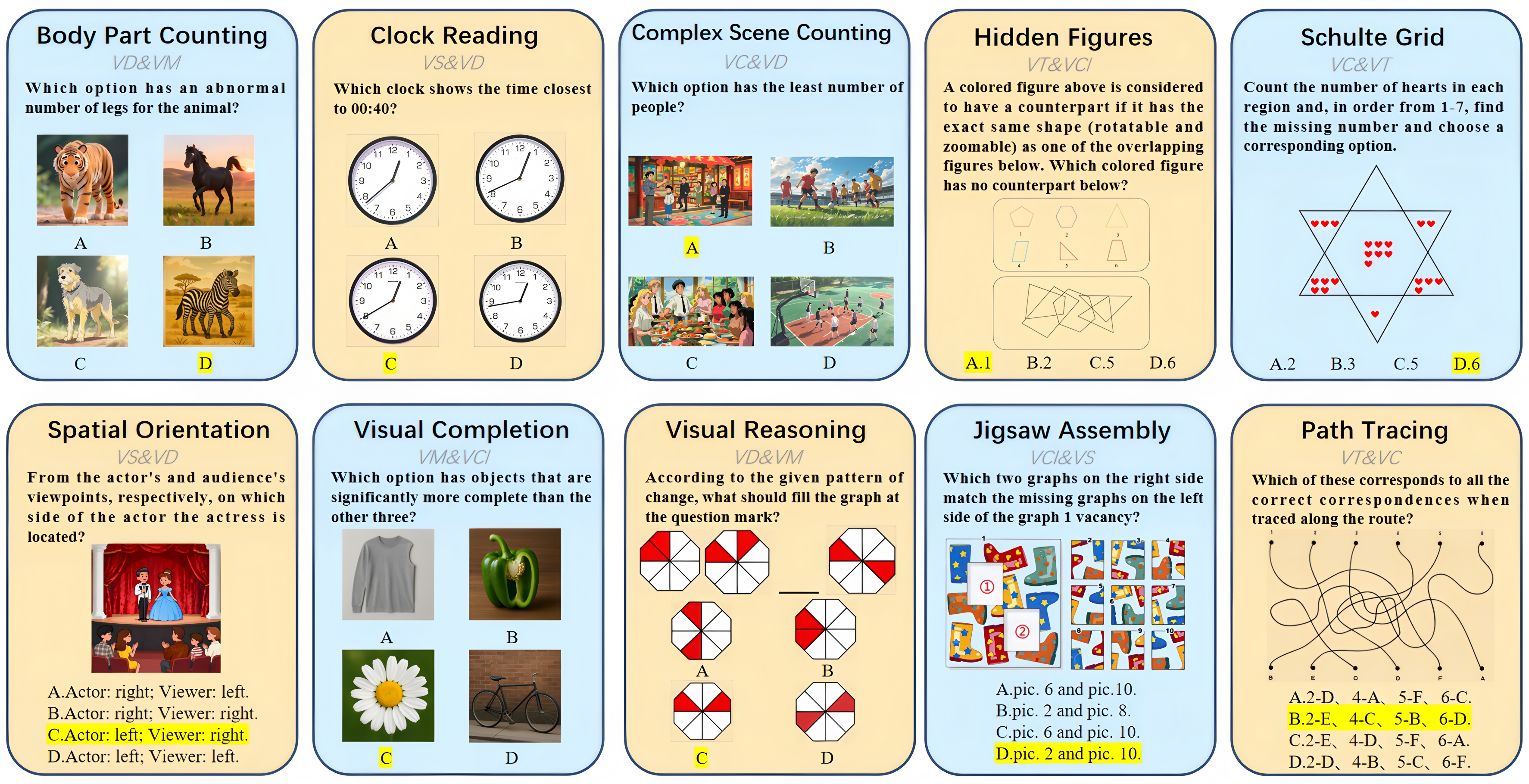}}
\caption{Task Instantiations of KidVis Benchmark.}
\label{fig02}
\end{figure*}

To ensure a comprehensive evaluation, we translate the six visual capabilities into ten distinct visual tasks. Each task contains 50 single-choice questions and employs a dual-factor design strategy, probing two primary visual primitives simultaneously, as shown in Fig.~\ref{fig02} and Table~\ref{tab01}. This approach reflects the complexity of ecological vision, where perceptual functions rarely operate in isolation. By combining these capabilities (e.g., Task 10 tests both Visual Tracking and Visual Concentration), KidVis prevents models from exploiting shortcuts and ensures that scores reflect strong, generalized visual intelligence rather than memorizing simplistic, single-dimensional problems. All question images within the KidVis benchmark are rendered at a high resolution of 2K ($2048 \times 2048$) or above to ensure that fine-grained details are preserved. Unlike benchmarks requiring encyclopedic knowledge, KidVis tasks are low-semantic-dependent and motor-free, minimizing confounding variables to isolate pure visual processing. Further details of the KidVis benchmark are presented in Appendix A.

\begin{itemize}
\item Task 1: Body Part Counting. This task fuses abnormal biological features (e.g., polydactyly) with naturalistic textures. Participants must separate specific features from background noise and contrast them against biological prototypes retrieved from long-term memory.
\item Task 2: Clock Reading. Participants must distinguish the hour and minute hands based on subtle length and width differences, and map their precise angular positions to the clock face reference frame to acquire temporal information.
\item Task 3: Complex Scene Counting. This task requires Figure-Background Separation under challenging conditions, such as occlusion or noise. Participants must sustain high visual concentration to filter out interference while employing fine-grained discrimination to identify targets based on local cues.
\item Task 4: Hidden Figures. Participants must track a specific geometric shape embedded within a complex network of intersecting lines. The Visual Closure mechanism is invoked to mentally complete the truncated segments, allowing participants to perceive the shape as a coherent Gestalt.
\item Task 5: Schulte Grid. Participants are required to searching for a strictly ascending numerical sequence (e.g., 1-16) within a scrambled grid. The participants must utilize Working Memory to maintain the search target while performing rigorous saccadic eye movements to locate the next target, inhibiting interference from neighboring digits.
\item Task 6: Spatial Orientation. Sub-tasks involve: (1) Determining relative object positions (e.g., coverage, rotation) within a reference frame; (2) 3D Structural Completion (calculating missing blocks); (3) Viewpoint transformation (e.g., identifying top-down views). Participants must construct 2D/3D coordinate systems to facilitate mental rotation and perspective taking.
\item Task 7: Visual Completion. Presented with objects containing structural defects, participants must retrieve the canonical prototype from memory. The cognitive system detects interrupted contours and performs Amodal Completion to instinctively "fill in" missing part, thereby identifying the complete or defective object.
\item Task 8: Visual Reasoning. Sub-tasks include pattern completion and Visual Sudoku. Participants are required to utilize visual discrimination to isolate features and detect state changes. Crucially, Visual Memory acts as a "Hypothesis Buffer," retaining previous states to infer the underlying rules of transformation.
\item Task 9: Jigsaw Assembly. Sub-tasks involve selecting fragments based on content compatibility, geometric decomposition, and boundary topology. Visual Closure acts as the generative prior, envisioning the complete state. Visual Spatial serves as the execution engine, manipulating fragments in a mental buffer to verify geometric fit and texture consistency.
\item Task 10: Path Tracing. Sub-tasks involve distinguishing entwined lines and following spiral mazes. Participants must execute "smooth pursuit" along thin paths, requiring high concentration to suppress interference and maintain continuity across long-range dependencies.
\end{itemize}

\section{Experiments}
% 定义图片中的颜色
\definecolor{humancolor}{HTML}{E6F4EA} % Human行的浅绿色
\definecolor{bestcolor}{HTML}{FCE4D6}   % SOTA结果的浅橙色
\definecolor{bestcolor_open}{HTML}{E1F5FE}   % open_source_SOTA结果的浅橙色

% 将表格字体设为 \scriptsize，并压缩行间距
\begin{table*}[t]

\caption{
  \parbox{\linewidth}{%
    \vspace{2mm}
    \normalfont 
    \raggedright 
    Results of mainstream MLLMs on six visual perceptual capabilities. The results of human children (6-7 years old) are marked with green; the highest scores of proprietary MLLMs are marked with orange; and the highest scores of open-source MLLMs are marked with blue.
  }
}
\label{tab02}

% 2. 压缩表格内部行间距 (默认是 1.0，改为 0.85 或 0.8 可大幅节省高度)
% \renewcommand{\arraystretch}{0.85} 

% 3. 使用 scriptsize 字体 (比 footnotesize 更小，适合大表格)
\scriptsize
\centering
\setlength{\tabcolsep}{5mm}
% 4. 这里的 resizebox 可以保留，但内容已经被我们手动压缩过了
% \resizebox{\textwidth}{!}{%
\begin{tabular}{l|c|cccccc}
\toprule
\textbf{Method} & \textbf{Mean} & \textbf{VC} & \textbf{VT} & \textbf{VD} & \textbf{VM} & \textbf{VS} & \textbf{VCl} \\
\midrule
 Human children & \cellcolor{humancolor}95.32 & \cellcolor{humancolor}95.11 & \cellcolor{humancolor}95.78 & \cellcolor{humancolor}95.47 & \cellcolor{humancolor}96.44 & \cellcolor{humancolor}93.78 & \cellcolor{humancolor}95.33 \\
\midrule
\multicolumn{8}{c}{\textit{Proprietary Multimodal Large Language Models}} \\
\midrule
GPT-4o & 40.05 & 34.67 & 28.67 & 41.60 & 52.67 & 35.33 & 47.33 \\
GPT-5 & \cellcolor{bestcolor}67.33 & 58.00 & 48.00 & \cellcolor{bestcolor}78.00 & \cellcolor{bestcolor}82.00 & \cellcolor{bestcolor}67.33 & \cellcolor{bestcolor}70.67 \\
Gemini-2.5-Flash & 56.71 & 49.33 & 39.33 & 63.60 & 75.33 & 52.67 & 60.00 \\
Gemini-2.5-Pro & 65.51 & \cellcolor{bestcolor}64.00 & \cellcolor{bestcolor}50.67 & 74.40 & 81.33 & 60.00 & 62.67 \\
\midrule
\multicolumn{8}{c}{\textit{Open-source Multimodal Large Language Models}} \\
\midrule
Qwen3-VL-4B-Instruct & 34.31 & 35.33 & 24.00 & 39.20 & 40.67 & 32.67 & 34.00 \\
Qwen3-VL-8B-Instruct & \cellcolor{bestcolor_open}38.58 & \cellcolor{bestcolor_open}43.33 & \cellcolor{bestcolor_open}32.67 & \cellcolor{bestcolor_open}40.80 & \cellcolor{bestcolor_open}42.00 & 34.00 & 38.67 \\
Qwen3-VL-30B-Instruct & 36.27 & 41.33 & 30.00 & 39.60 & 38.67 & 33.33 & 34.67 \\
InternVL3\_5-4B & 31.69 & 35.33 & 29.33 & 30.80 & 29.33 & 33.33 & 32.00 \\
InternVL3\_5-8B & 30.27 & 31.33 & 26.67 & 29.60 & 30.67 & 30.00 & 33.33 \\
InternVL3\_5-14B & 34.13 & 40.00 & 28.00 & 38.80 & 31.33 & \cellcolor{bestcolor_open}36.00 & 30.67 \\
InternVL3\_5-38B & 33.82 & 40.67 & 30.00 & 29.60 & 30.67 & 32.00 & \cellcolor{bestcolor_open}40.00 \\
llava-next-8b-hf & 23.78 & 17.33 & 18.00 & 28.00 & 29.33 & 24.67 & 25.33 \\
llava-next-72b-hf & 30.09 & 28.00 & 21.33 & 33.20 & 41.33 & 24.00 & 32.67 \\
llava-next-110b-hf & 29.78 & 30.67 & 26.67 & 30.00 & 32.67 & 26.67 & 32.00 \\
llava-onevision-qwen2-7b-ov-hf & 29.42 & 34.00 & 24.67 & 31.20 & 32.67 & 24.00 & 30.00 \\
llava-onevision-qwen2-72b-ov-hf & 33.64 & 35.33 & 27.33 & 33.20 & 36.00 & 30.00 &\cellcolor{bestcolor_open} 40.00 \\
gemma-3-27b-it & 33.07 & 38.00 & 28.00 & 34.40 & 32.67 & 30.67 & 34.67 \\
Kimi-VL-A3B-Instruct & 34.89 & 38.67 & 25.33 & 38.00 & 39.33 & 33.33 & 34.67 \\
mPLUG-owl3-2B & 24.84 & 24.00 & 26.00 & 24.40 & 24.00 & 27.33 & 23.33 \\
mPLUG-owl3-7B & 25.42 & 23.33 & 20.00 & 27.20 & 30.00 & 25.33 & 26.67 \\
\bottomrule
\end{tabular}%
% }

% 5. 在表格结束后，添加负的垂直间距，把下方的正文“拉”上来
\vspace{-4mm} 
\end{table*}

\subsection{Experimental Setup}

\paragraph{Multimodal Large Language Models}Based on KidVis benchmark,we evaluate 20 state-of-the-art MLLMs , spanning both proprietary and open-source models. The suite includes leading proprietary MLLMs (e.g., GPT-5, Gemini-2.5-Pro) and comprehensive open-source MLLMs (e.g., Qwen3-VL, InternVL3.5, LLaVA-Next), covering parameter scales from 2B to 110B. All models are evaluated in a zero-shot setting using default configurations to assess their visual perceptual capabilities without task-specific tuning.
\paragraph{Human Physiological Baseline}To establish a rigorous physiological upper bound, we recruited three children aged 6-7 years. This age group, characterized by largely matured visual cortices yet minimal encyclopedic semantic interference\cite{ref30}, serves as the ideal benchmark for evaluating pure visual perceptual capabilities.
\paragraph{Evaluation Metrics}We employ a two-tiered metric system to quantify performance:
Task Score (0-100): This is calculated strictly based on the correctness of the final option. For each of the 50 questions per task, participants award 2 points for a correct response and 0 points for an incorrect one, resulting in a normalized 100-point scale.
Capability Score(0-100): This derives from the weighted aggregation of tasks probing the same visual perception 
perceptual capabilities. For instance, the capability score for Visual Tracking is computed as the weighted average of results from Task 4, Task 5, and Task 10.

\subsection{Main Results}

\begin{figure*}[htbp]
\centerline{\includegraphics[width=0.8\textwidth]{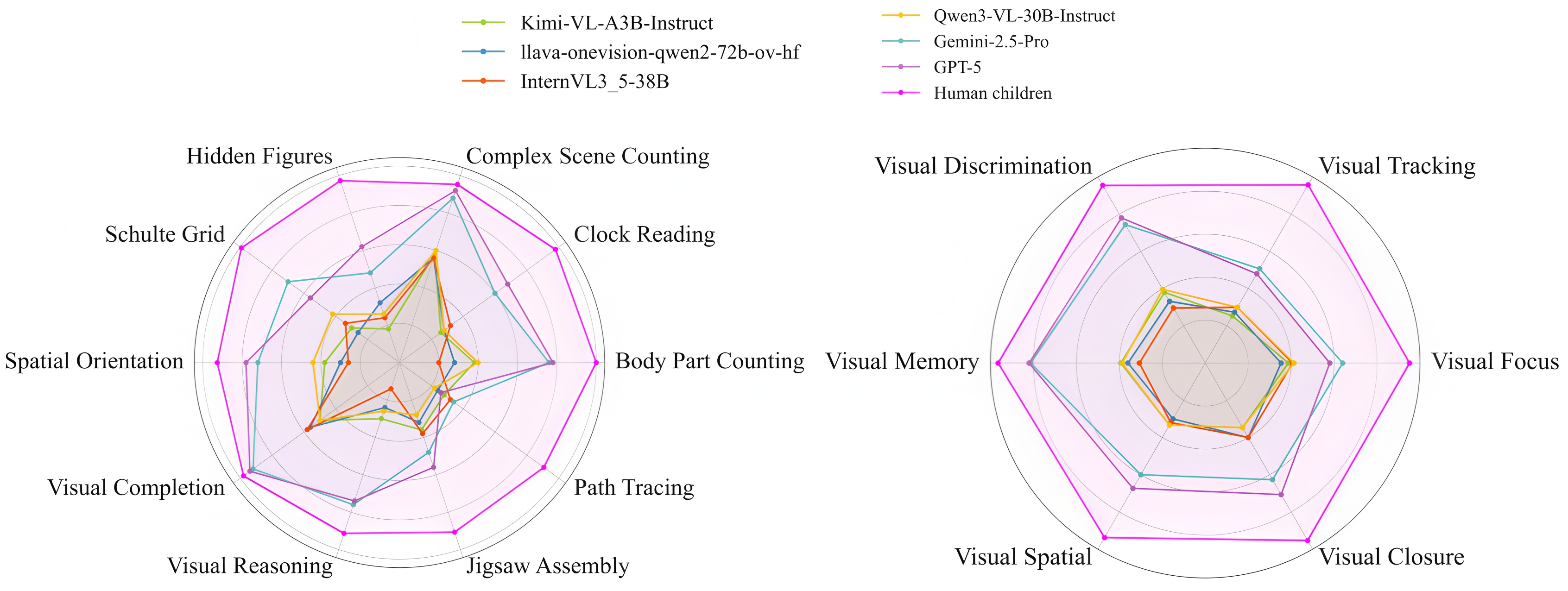}}
\caption{Comparison of representative MLLMs on KidVis benchmark. Left: Performance on 10 tasks within KidVis benchmark. Right: Performance on six visual perception capabilities.}
\label{fig03}
\end{figure*}

\begin{figure*}[htbp]
\centerline{\includegraphics[width=\textwidth]{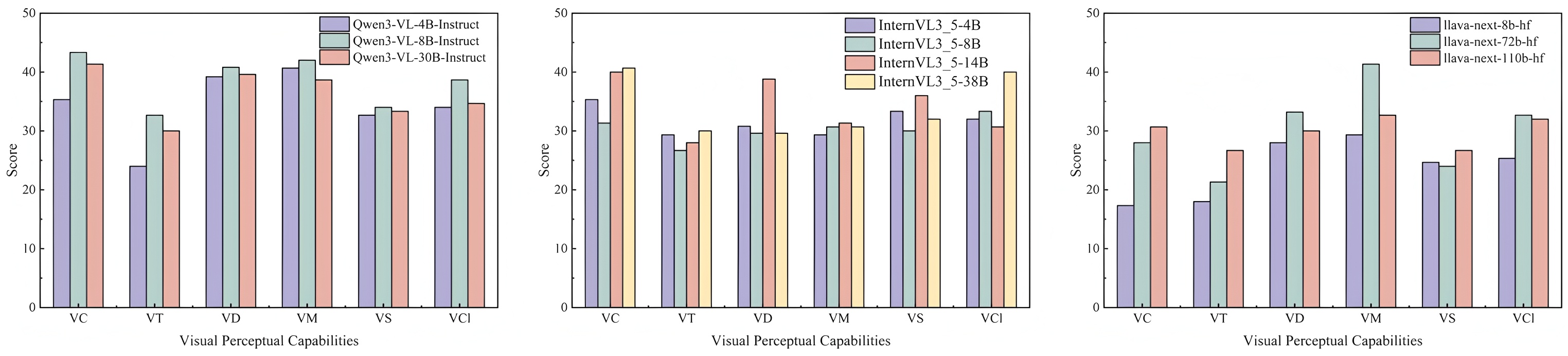}}
\caption{Impact of open-source MLLMs parameters across three major families (Qwen3-VL, InternVL3.5, LLaVA-Next).}
\label{fig04}
\end{figure*}

The quantitative results on our KidVis benchmark are shown in Table~\ref{tab02} and Fig.~\ref{fig03}, which reveal a profound separation between semantic reasoning and visual perception. Further details of the experimental results are shown in Appendix B.

\paragraph{Human Ceiling}The human group demonstrates near-saturation performance, achieving a composite score of 95.32 across all capabilities. This confirms that KidVis tasks, while challenging for machines, are instinctive to the human visual system.
\paragraph{Proprietary MLLMs Gap}In stark contrast, a significant quantitative gap persists between proprietary models and the human baseline. The state-of-the-art model GPT-5 attains only 67.33 across all capabilities, falling behind the human score of 95.32 by nearly 28 percentage points. Similarly, Gemini-2.5-Pro exhibits a comparable gap, scoring 65.51, which further confirms this trend. Although these proprietary models outperform their open-source counterparts by a large margin, the data reveals a systemic performance gap of around 30 percentage points separating even the most advanced MLLMs from the physiological baseline of a 6-year-old child.
\paragraph{Open-Source MLLMs Collapse}The performance degradation is catastrophic for open-source models. Leading series such as Qwen3-VL and InternVL3.5 fall behind the human baseline by over 50\%, with average scores typically falling below 40. In tasks requiring precise spatial manipulation (e.g., Visual Tracking), many of these models achieve performance indistinguishable from random chance.

\section{Analysis}

\subsection{Analysis of Atomic Capabilities}
We conduct an analysis of the models’ performance on the six atomic capabilities to facilitate a more comprehensive horizontal and vertical examination.
\paragraph{Visual Concentration. Soft vs. Hard Attention}Humans employ hard attention (fixation) to physically exclude surrounding stimuli. In contrast, MLLMs rely on soft attention, assigning non-zero weights to all visual tokens. In high-interference tasks like the Schulte Grid, this global receptivity causes noise accumulation and attentional drift, as the models lack an inherent inhibition mechanism to actively suppress background distractors.
\paragraph{Visual Tracking. Discontinuity}Visual tracking is the visual perceptual ability in which MLLMs exhibit the worst performance. We suggest that Standard Vision Transformer (ViT) architectures, which employ patch-based tokenization to divide continuous images into discrete grids, may fundamentally disrupt continuous lines\cite{ref33,ref34,ref35}. Consequently, models fail to maintain connectivity constraints in tasks like Path Tracing, statistically jumping to intersecting lines with similar textures rather than tracing the actual path.
\paragraph{Visual Discrimination. Semantic Prior Bias and Neglect of Fine-grained Visual Cues}While proprietary MLLMs like GPT-5 demonstrate robust capabilities (scoring 78.00), open-source models significantly fall behind. We found two distinct failure mechanisms. First, MLLMs exhibit a severe Semantic Prior Bias. In tasks like Fine-grained Body Part Counting, when visual evidence (e.g., a five-legged cat) conflicts with pre-trained knowledge, models frequently hallucinate biologically correct but visually false answers, confirming that the concept is superior to the perception. Second, models suffer from fundamental Perceptual Imprecision. This is starkly illustrated in Task 3 (e.g., identifying pentagons). Although models correctly ground their search in the semantic criteria (i.e., 5-sided closed polygons), they lack the visual acuity to strictly differentiate these specific contours from visually similar distractions.
\paragraph{Visual Memory. Operational Rigidity}Visual memory stands as the strongest capability for proprietary models, with GPT-5 achieving a high score of 82.00. This confirms that top-tier MLLMs possess robust static contextual storage. However, open-source models fall behind significantly, with the majority scoring below 40. This stark gap suggests that the bottleneck lies not in raw storage capacity, but in operational fluency. Unlike humans, who utilize visual memory as a dynamic read-write workspace for mental simulation (e.g., rotating objects in Visual Reasoning), MLLMs may treat visual history as a static read-only buffer, failing to execute the active state transitions required for sequential inference.
\paragraph{Visual Spatial. Reference Frame Loss}Analysis of Spatial Orientation reveals a critical failure in processing depth cues like perspective, where models struggle to infer coherent 3D scenes from 2D images. This points to a fundamental absence of an internal reference frame. Without a grounded coordinate system, models cannot execute viewpoint transformation, leading to an inability to distinguish between inherent object features and viewpoint-dependent states (e.g., the relative relationship of left and right). Instead, they rely on coarse semantic approximation , as seen in Clock Reading (e.g., "pointing near 7") , lacking the precise metric computation required for strict spatial reasoning.
\paragraph{Visual Closure. Texture-Geometry Separation}Our analysis of Jigsaw Assembly exposes a fundamental Texture-Geometry Separation. While humans use the Gestalt Law of Continuity to mentally reconstruct the spatial geometry of missing parts—matching them via characteristic edge outlines or internal details—MLLMs prioritize superficial texture consistency (e.g., similar colors or patterns). Consequently, models struggle to visually trace connecting lines, failing to strictly verify whether the boundary and interior textures are truly compatible.

\subsection{The Scaling Law Paradox}

A prevailing principle in modern Artificial Intelligence is the Scaling Law—the expectation that performance improves consistently with model size. However, our evaluation across three major open-source families (Qwen3-VL, InternVL3.5, LLaVA-Next) reveals a startling exception: physiological visual primitives do not scale linearly with parameter count.
As illustrated in Fig.~\ref{fig04}, the Qwen3-VL series aggregate performance peaks at the 8B scale and regresses at 30B. We suggest that this stems from Semantic Interference: as the model scales, the linguistic decoder becomes excessively powerful and over-reasons simple perceptual tasks, generating semantically reasonable but visually ungrounded details. The InternVL3.5 and LLaVA-Next series demonstrate performance saturation, where substantial parameter scaling yields tiny gains in visual perception. This evidence highlights that although reasoning capabilities benefit from scale effects, basic visual perception appears to demand architectural innovation.

\section{conclusion}
By establishing the KidVis benchmark, this study systematically quantifies the performance of existing MLLMs across six fundamental visual capabilities, shifting the evaluation paradigm from expert-level reasoning to child-level perception. Our results demonstrate that even state-of-the-art models exhibit significant deficits when confronted with low-semantic visual tasks that children perform instinctively. Crucially, we uncover a Scaling Law Paradox: simply increasing model parameters is ineffective for enhancing, and may even weaken, innate visual abilities. These findings offer critical insights for future research, suggesting that equipping models with true physical perception requires simulating human visual developmental processes and architectural innovation.

\bibliographystyle{IEEEbib}
\bibliography{icme2026_template_anonymized}

\vspace{12pt}

\end{document}